# Research on Optimizing Real-Time Data Processing in High-Frequency Trading Algorithms using Machine Learning

Yuxin Fan[*]
School of Engineering and Applied Science
University of Pennsylvania
Canada, Toronto
[*] Corresponding e-mail address: yuxinfan@alumni.upenn.edu

Lei Fu
School of Computer Science and Engineering
Changshu Institute of Technology
Suzhou, China
E-mail address: fuleiac@gmail.com

Liyang Wang
Olin Business School,
Washington University in St. Louis
St. Louis, MO, 63130, USA
E-mail address: liyang.wang@wustl.edu

Zhuohuan Hu
Independent Researcher
San Francisco USA
E-mail address: ZHUOHUAN.tommy@gmail.com

Yu Cheng
Independent Researcher
Mclean, VA, 22102, USA
E-mail address: yucheng576@gmail.com

Yuxiang Wang
Independent Researcher
New York,NY,USA
E-mail address: yuxiang.wang0476@gmail.com

*Abstract*—High-frequency trading (HFT) represents a pivotal and intensely competitive domain within the financial markets. The velocity and accuracy of data processing exert a direct influence on profitability, underscoring the significance of this field. The objective of this work is to optimise the real-time processing of data in high-frequency trading algorithms. The dynamic feature selection mechanism is responsible for monitoring and analysing market data in real time through clustering and feature weight analysis, with the objective of automatically selecting the most relevant features. This process employs an adaptive feature extraction method, which enables the system to respond and adjust its feature set in a timely manner when the data input changes, thus ensuring the efficient utilisation of data. The lightweight neural networks are designed in a modular fashion, comprising fast convolutional layers and pruning techniques that facilitate the expeditious completion of data processing and output prediction. In contrast to conventional deep learning models, the neural network architecture has been specifically designed to minimise the number of parameters and computational complexity, thereby markedly reducing the inference time. The experimental results demonstrate that the model is capable of maintaining consistent performance in the context of varying market conditions, thereby illustrating its advantages in terms of processing speed and revenue enhancement.

*Keywords-High-Frequency Trading, Real-Time Data Processing, Dynamic Feature Selection, Lightweight Neural Network.*

## I. Introduction

High-frequency trading (HFT) has become a fundamental aspect of contemporary financial markets, with a considerable number of trades being executed at intervals of less than a millisecond. This type of trading fulfils a number of roles within the market, including the provision of liquidity and the capture of minor price movements, thereby assisting the market in achieving greater efficiency. However, this rapid and complex trading environment places significant demands on the real-time data processing capabilities of financial institutions. The ability to process data rapidly and efficiently is crucial for success in this environment, where latency can have a significant impact on the outcome of trades [1].

Latency represents an opportunity cost, with even a few microseconds of lag potentially leading to missed or lost trading opportunities. Consequently, market participants are continually striving to enhance the velocity and responsiveness of their trading systems, investing in state-of-the-art hardware, network infrastructure, and sophisticated software solutions. While traditional algorithmic methods have demonstrated excellent performance over the past few decades, particularly in technical analysis and quantitative trading strategies, they are often inadequate to meet the demands of modern high-frequency trading, given the increasing complexity and competitiveness of modern financial markets. When confronted with abrupt and erratic market movements, conventional techniques may prove inadequate in adapting strategies in a timely manner, due to their inherent lack of flexibility [2].

As a potent instrument for surmounting these obstacles, machine learning furnishes data-driven methodologies that can discern and adapt to intricate market behaviours. By leveraging vast quantities of historical data and real-time data streams, machine learning algorithms are capable of identifying hitherto unrecognised market patterns, formulating predictions and acting upon them in a matter of seconds. Nevertheless, the application of machine learning to high-frequency trading presents a distinct set of challenges [3]. Firstly, the construction of ultra-low latency machine learning models presents a significant challenge, as sophisticated deep learning algorithms, while offering high levels of accuracy, typically necessitate the utilisation of substantial computing resources, which can consequently result in processing delays. Secondly, the challenge is to achieve an equilibrium between model complexity and processing speed [4]. This is crucial to ensure that the model is sufficiently robust to navigate complex markets and maintain efficient operation in a rapidly evolving environment. Furthermore, the real-time data traffic inherent to high-frequency trading is both considerable and rapidly evolving. Consequently, machine learning models must demonstrate the capacity to maintain consistent performance and respond expeditiously despite the precipitous alterations in data inputs.

The confluence of machine learning and real-time processing in high-frequency trading necessitates the development of an innovative framework that can address these distinctive requirements. Such a framework should be highly adaptive, capable of dynamically selecting and adjusting model input variables to reduce noise and improve prediction accuracy [5]. Concurrently, the framework must be designed to be efficient and lightweight, ensuring minimal latency in data processing and decision-making. Only when these conditions are met can machine learning methods genuinely enhance the performance of high-frequency trading systems and confer a competitive advantage upon traders.

## II. RELATED WORK

In a pioneering contribution, Passalis et al. [6] advanced an innovative approach that integrates deep learning with adaptive data normalisation techniques, thereby enabling the model to perform data normalisation in a dynamic and adaptive manner during the training process. This method markedly enhances the resilience and adaptability of the model when confronted with non-stationary data or anomalous data, and effectively addresses the instability of traditional models in response to alterations in data distribution. Fu Y et al. [7] integrated volatility estimation into a functional time series model, thereby constructing a complex model for stock index futures forecasting. The incorporation of volatility estimation into the model enhances its capacity to discern potential nonlinearities and non-stationarities in the data, thereby enhancing the accuracy of the predictions. Liu et al. [8] put forth a nonstationary time series forecasting method based on Transformer architecture. This method eliminates the adverse impact of nonstationary characteristics on the prediction model by transforming the original data in the preprocessing stage and restoring it in the subsequent stage.

In a previous study, Li et al. [9] proposed a dynamic feature selection algorithm based on the Q-learning mechanism. This algorithm is designed to select the most valuable features based on the weight of the current features. During the training process, the algorithm continually modifies the weights and selection strategies of features in accordance with the classifier's performance on the validation set, thereby optimising the classifier's overall performance. The experimental results demonstrate the efficacy of this method in improving classification performance and addressing complex data feature selection issues, particularly in high-dimensional datasets. Ali et al. [10] put forth an evolutionary metric filter method based on reinforcement learning. The algorithm incorporates evolutionary algorithms for population generation and modifies the metric function of each individual through reinforcement learning, thereby optimising the feature selection process.

## III. METHODOLOGIES

### A. Dynamic Feature Selection

The market data should be allowed to flow freely $X = \{x_1, x_2, \ldots, x_n\}$ represents the set of all data vectors $x_i$, each of which belongs to the $d$-dimensional feature space $\mathbb{R}^d$. The cluster centre $\mu_j$ is initially determined using the improved $k-$means$++$ clustering method, expressed as Equation 1.

$$\mu_j = arg \min_{\mu} \sum_{i=1}^{n} \min_{j} \parallel x_i - \mu_j \parallel^2 . \#(1)$$

The significance of each feature is quantified by the weighted variance, $\sigma_{i,j}^2$, as illustrated in Equation 2.

$$\sigma_{i,j}^2 = \frac{1}{|C_j|} \sum_{x \in C_j} (x_i - \mu_{i,j})^2, \#(2)$$

where $\mu_{i,j}$ represents the mean value of feature $i$ within cluster $C_j$. The feature weight, $w_i$, is then calculated as the reciprocal of the weighted mean variance across all clusters. This approach emphasises the priority of features with low variance, as illustrated in Equation 3.

$$w_i = \left( \sum_{j=1}^{k} p(C_j) \cdot \sigma_{i,j}^2 \right)^{-1}, \#(3)$$

where $p(C_j) = \frac{|C_j|}{n}$ represents the probability of cluster $C_j$.

In order to identify the most pertinent features, we employ a process of maximising mutual information $I(x_i; y)$, which enables us to ascertain the degree of relevance of a given feature to the predicted target. The mutual information can be calculated using the following Equation 4.

$$I(x_i; y) = \sum_{x_i} \sum_{y} p(x_i, y) log \frac{p(x_i, y)}{p(x_i)p(y)}, \#(4)$$

where $p(x_i, y)$ represents the joint probability distribution, while $p(x_i)$ and $p(y)$ denote the marginal probability distributions. In high-frequency trading, the estimation of the joint distribution must be updated in real time. This is achieved through the use of the Kernel Density Estimation (KDE) method, as illustrated in Equation 5.

$$p(x_i, y) = \frac{1}{nh} \sum_{t=1}^{n} K(\frac{x_i - x_i^{(t)}}{h}) K(\frac{y - y^{(t)}}{h}), \#(5)$$

where the symbol $K(\cdot)$ represents the kernel function, while $h$ denotes the bandwidth parameter.

*B. Lightweight Neural Networks*

In the design of lightweight neural networks, the output $Z$ of the convolutional layer is defined by the following Equation 6.

$$Z_{m,n} = \sigma\left(\sum_{i=1}^{f}\sum_{j=1}^{f} W_{i,j} \cdot X_{m+i-1,n+j-1} + b\right), \#(6)$$

In this context, the variable $f$ represents the size of the convolution kernel, $W_{i,j}$ denotes the weight matrix associated with the convolution kernel, $\sum_{j=1}^{f} W_{i,j} \cdot X_{m+i-1,n+j-1} + b$ is the bias term, and $\sigma(\cdot)$ signifies a nonlinear activation function, such as the rectified linear unit function.

In order to further reduce the computational complexity, pruning techniques are employed in order to select the most important parameters. In accordance with the specified threshold of $\epsilon$, the set of parameters $W_{pruned}$ designated for pruning is represented by the following Equation 7.

$$W_{pruned} = \{w_{i,j} \in W \mid |w_{i,j}| > \epsilon\}. \#(7)$$

In order to achieve a balance between model performance and parameter reduction, Lagrangian optimisation was employed during the pruning process. The objective function is presented in Equation 8.

$$\mathcal{L}(W, \lambda) = Loss(W) + \lambda \sum_{i=1}^{d} \mathbb{I}(|w_i| \leq \epsilon), \#(8)$$

where $Loss(W)$ represents the training error, $\mathbb{I}(\cdot)$ is the indication function, and $\lambda$ is the regularisation parameter, which controls the intensity of pruning.

The network comprises L convolution modules, each of which employs a distinct convolutional kernel size F for the extraction of multi-scale features. The output fusion is weighted and summed, as illustrated by Equation 9.

$$\hat{y} = \sum_{i=1}^{l} \alpha_i \cdot M_i(X), \#(9)$$

The optimisation of $\alpha_i$ is achieved through backpropagation, which allows for the adjustment of the contribution of each module to the final output.

## IV. EXPERIMENTS

*A. Experimental Setups*

We selected 50 ETF datasets for model validation and evaluation. The 50ETF dataset contains the historical trading data of the 50 ETF on the Shanghai Stock Exchange, including the opening price, closing price, highest price, lowest price, trading volume and other characteristics. These datasets can reflect the volatility and trend changes of the market, providing rich characteristic information for high-frequency trading predictions. The dynamic feature selection module uses the k-means clustering algorithm, the number of clusters k is set to 10, and the feature weights are adjusted once in each training cycle. In a lightweight neural network, the kernel sizes of the convolutional layer are 3×3 and 5×5, the pruning threshold $\varepsilon$ is set to 0.01, and the regularization parameter $\lambda$ is 0.1. The model uses the Adam optimizer, with an initial learning rate of 0.001, decaying to half of its original size every 10 training cycles, and a batch size of 128, ensuring efficient training and stable convergence on large-scale data.

*B. Experimental Analysis*

In the experimental comparison, a number of benchmark methods were selected for analysis. The DL-AN method employs a combination of deep learning and adaptive normalization techniques to enhance the resilience of the model in the presence of non-stationary and anomalous data. The VE-FTS method integrates volatility estimation into time series models, thereby enhancing the accuracy of predictions for nonlinear and non-stationary data. Through data preprocessing and restoration, T-TS mitigates the impact of non-stationarity on prediction. DF-QA performs dynamic feature selection based on Q learning, thereby enhancing the performance of high-dimensional data classification. RL-EMF combines evolutionary algorithms and reinforcement learning to optimise feature selection, and utilises "statistical metric functions" to enhance the reliability of feature evaluation.

The autocorrelation function (ACF) is a statistical tool employed to assess the correlation between distinct lags in time series data. Figure 1 illustrates the findings of the ACF analysis of the trading price factor. It demonstrates that the correlation does not exhibit a gradual decay with the lag period, but rather a fluctuating state.

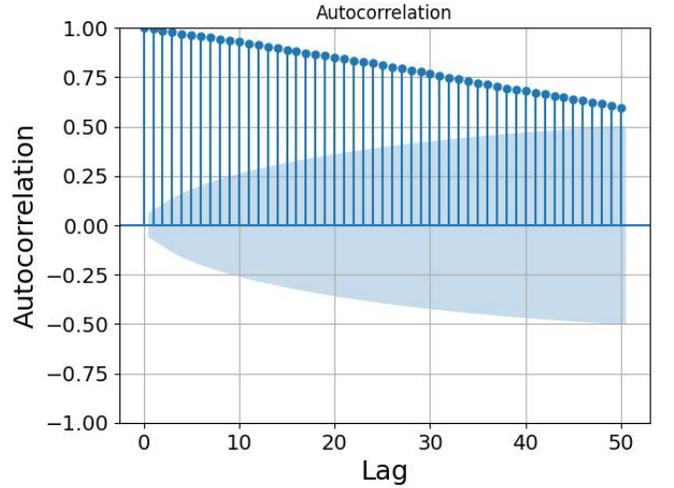

Figure 1. ACF Analysis of Trading Price Factor.

This indicates the presence of a notable correlation between the price factor in the high-frequency trading data and its own distinct lags in time, which may suggest a certain degree of dependence or a repeating pattern of price movements in the short term. The application of ACF analysis facilitates the identification of autocorrelation structures within time series data, thereby providing a foundation for the selection and construction of models.

Figure 2 shows the comparison of the root mean square error of different methods, reflecting the prediction performance of the model with different parameter settings by increasing the size of the data window. It can be seen that the Ours method exhibits the lowest RMSE across all parameters, indicating its superiority in prediction accuracy and stability. Other methods, such as DL-AN and T-TS, although performing better at certain window sizes, have greater overall volatility and higher RMSE. The error range shading shows the stability of the model under different parameters, and the Ours method has the least fluctuation at multiple parameter values, indicating that it is more robust and consistent when dealing with different data windows.

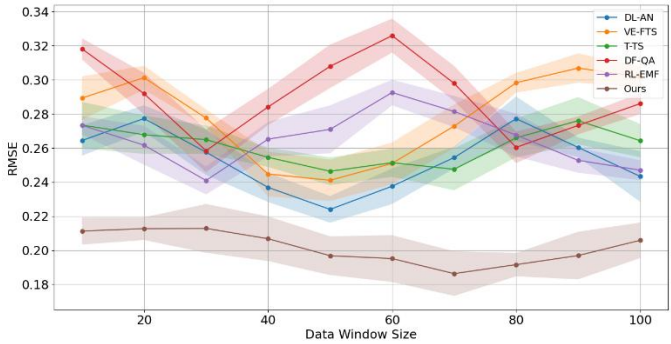

Figure 2. Comparison of RMSE for Different Methods.

Figure 3 illustrates the R² accuracy of diverse methodologies with varying mutual information thresholds. Additionally, error shading has been incorporated to demonstrate the fluctuation range of the results. The mutual information threshold is employed for the purpose of identifying salient features, which exert a significant influence on the predictive performance of the model. The R²-precision index gauges the model's capacity to elucidate data variability. A value approaching 1 signifies a higher degree of accuracy in predicting the outcome. As illustrated in Figure 3, the Ours method demonstrates consistent and high R² accuracy across a range of thresholds, exhibiting stability within the error range.

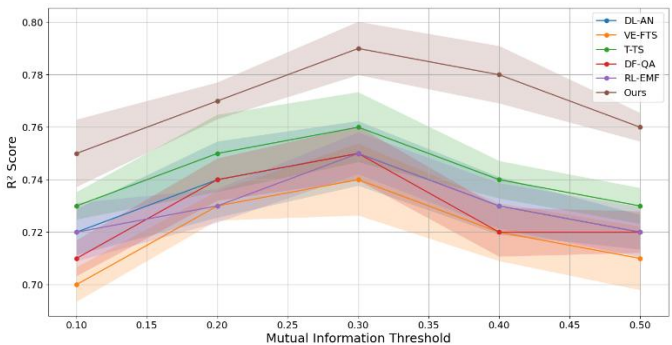

Figure 3. Mutual Information Threshold vs R² Score.

This suggests that it achieves an optimal balance between feature selection and accuracy improvement. In contrast, the performance of other methods is observed to fluctuate, thereby demonstrating the adaptability of the model under different feature selection strategies.

Table 1 presents a comparative analysis of the various methodologies in terms of their average execution time and delay. The Average Execution Time metric quantifies the time elapsed between the reception of input and the completion of a computation. As can be observed, the Ours method demonstrates the most optimal performance in both metrics, exhibiting the lowest average execution time and latency of 35 ms and 10 ms, respectively. This illustrates its notable advantage in high-frequency trading and real-time forecasting scenarios. Other methods, such as DL-AN and VE-FTS, exhibit relatively high execution times and delays, indicating that they are somewhat less efficient than Ours when processing rapidly evolving data.

TABLE I. AVERAGE EXECUTION AND LATENCY COMPARISON RESULTS

| Method | Average Execution Time (ms) | Average Latency (ms) |
|---|---|---|
| DL-AN | 45 | 15 |
| VE-FTS | 50 | 18 |
| T-TS | 42 | 14 |
| DF-QA | 55 | 20 |
| RL-EMF | 48 | 17 |
| Ours | **35** | **10** |

## V. Conclusions

In conclusion, this study corroborates the benefits of the Ours method in terms of accuracy, execution time and delay through experimental comparison, and emphasises its exemplary performance in high-frequency trading. Further research could concentrate on optimising the complexity and scalability of the models, and investigate the combination with techniques such as graph neural networks or reinforcement learning. Additionally, there is scope for expansion of applications in areas such as risk prediction and portfolio optimisation.